\documentclass[sigconf,nonacm,balance=false]{acmart}
\AtBeginDocument{%
  }

\usepackage{booktabs}
\usepackage{array}
\usepackage{tabularx}
\usepackage{xcolor}
\usepackage{colortbl}
\usepackage{multirow}
\usepackage{graphicx}
\usepackage{microtype}
\usepackage[most]{tcolorbox}
\definecolor{analysisfg}{HTML}{1746B5}
\definecolor{analysisbg}{HTML}{ECF2FF}
\definecolor{planfg}{HTML}{7540A0}
\definecolor{planbg}{HTML}{F4EEFA}
\definecolor{executefg}{HTML}{008543}
\definecolor{executebg}{HTML}{EAF7EF}
\definecolor{answerfg}{HTML}{C62828}
\definecolor{answerbg}{HTML}{FFF0F0}
\newcommand{\routertag}[3]{%
  \begingroup
  \setlength{\fboxsep}{0.8pt}%
  \colorbox{#2}{%
    \textcolor{#3}{\texttt{\textless#1\textgreater}}%
  }%
  \endgroup
}
\newcommand{\AnalysisTag}{%
  \routertag{analysis}{analysisbg}{analysisfg}}
\newcommand{\PlanTag}{%
  \routertag{plan}{planbg}{planfg}}

\newcommand{\AnswerTag}{%
  \routertag{answer}{answerbg}{answerfg}}

\makeatletter
\renewcommand{\@parfont}{\bfseries}
\makeatother

\makeatletter
\g@addto@macro\@adjustvertspacing{%
  \advance\abovedisplayskip 0pt minus 2pt\relax
  \advance\belowdisplayskip 0pt minus 2pt\relax}
\makeatother
\AtBeginDocument{\normalsize}
\setcopyright{none}
\begin{document}

\title{EvoRouter: Cost-Efficient Multi-Agent Routing via Insight-Guided On-Policy Distillation}

\author{Yitao Xiao}
\authornote{Equal contribution.}
\affiliation{%
  \institution{Beijing University of Posts and Telecommunications}
  \city{Beijing}
  \country{China}}
\email{yitooooom@bupt.edu.cn}

\author{Xu Zhang}
\authornotemark[1]
\affiliation{%
  \institution{Beijing University of Posts and Telecommunications}
  \city{Beijing}
  \country{China}}
\email{zhangxu2024@bupt.edu.cn}

\author{Ye Cheng}
\authornote{Corresponding authors.}
\affiliation{%
  \institution{Beijing University of Posts and Telecommunications}
  \city{Beijing}
  \country{China}}
\email{yecheng@bupt.edu.cn}

\author{Shaoyong Guo}
\affiliation{%
  \institution{Beijing University of Posts and Telecommunications}
  \city{Beijing}
  \country{China}}
\email{syguo@bupt.edu.cn}

\author{Yan Sun}
\affiliation{%
  \institution{Beijing University of Posts and Telecommunications}
  \city{Beijing}
  \country{China}}
\email{sunyan79@bupt.edu.cn}

\author{Ziyue Zhu}
\affiliation{%
  \institution{Beijing University of Posts and Telecommunications}
  \city{Beijing}
  \country{China}}
\email{zyzhu@bupt.edu.cn}

\author{Sitao Xie}
\affiliation{%
  \institution{Fudan University, Shanghai Innovation Institute}
  \city{Shanghai}
  \country{China}}
\email{26113050151@m.fudan.edu.cn}

\author{Guo Tong}
\affiliation{%
  \institution{Nanyang Technological University}
  \country{Singapore}}
\email{tong.guo@ntu.edu.sg}

\author{Haoran Luo}
\authornotemark[2]
\affiliation{%
  \institution{Nanyang Technological University}
  \country{Singapore}}
\email{haoran.luo@ntu.edu.sg}

\renewcommand{\shortauthors}{First Author et al.}

\begin{abstract}
  Serving web queries at scale with multi-agent systems requires jointly configuring large language models, collaboration modes, agent counts, interaction rounds, and skills. However, the resulting configuration space grows combinatorially, while sparse execution rewards provide limited fine-grained supervision for interdependent planning decisions. We propose \textbf{EvoRouter}, a self-evolving multi-agent routing framework driven by insight-guided on-policy distillation. A dedicated expert periodically synthesizes performance–cost feedback from configuration trajectories across tasks into reusable conditional routing insights about capability requirements, returns from resource scaling, and interactions among configuration choices. A privileged teacher combines these evolving insights with current execution feedback to provide dense token-level supervision for the router’s own planning trajectories through on-policy distillation. Cost-aware reinforcement learning jointly optimizes task performance and computational cost. Together, these training signals internalize execution insights into the routing policy. The updated policy generates new trajectories that support subsequent insight extraction, forming an execution-driven self-improvement loop. We evaluate EvoRouter on 19 in-domain datasets spanning five domains and two out-of-domain benchmarks. EvoRouter achieves an average in-domain accuracy of 68.69\%, outperforming the strongest routing baseline by 7.23 percentage points while reducing inference compute cost by 50.95\%. Our code is open-sourced at \url{https://github.com/CenYangtze/EvoRouter}.
\end{abstract}


\maketitle
\vspace{-6pt}

\section{Introduction}

\begin{figure}[!t]
  \centering
  \includegraphics[width=0.92\linewidth]{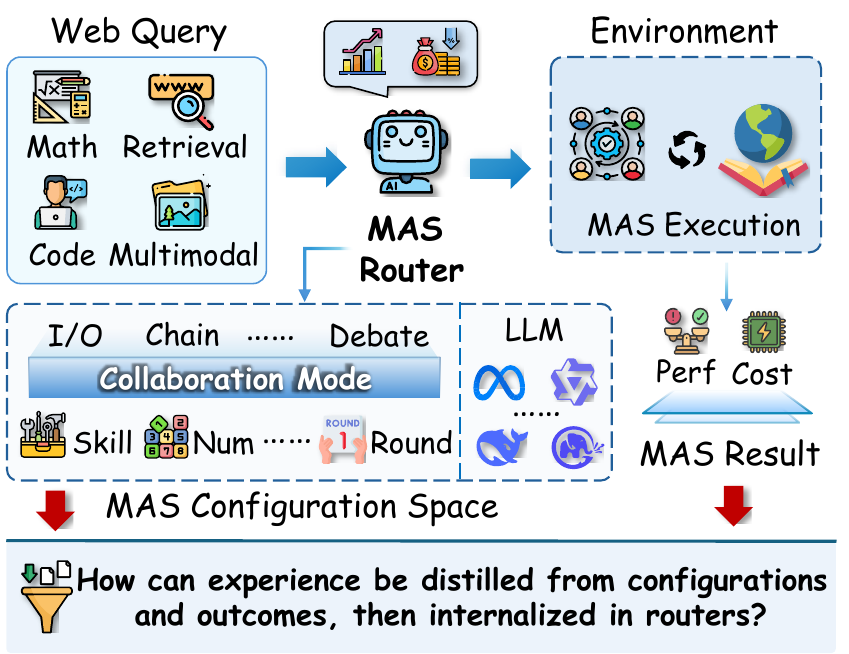}
  \caption{MAS router for web queries.}
  \label{fig:challenge}
  \vspace{-10pt}
\end{figure}

\begin{figure*}[t]
  \centering
  \includegraphics[width=\textwidth]{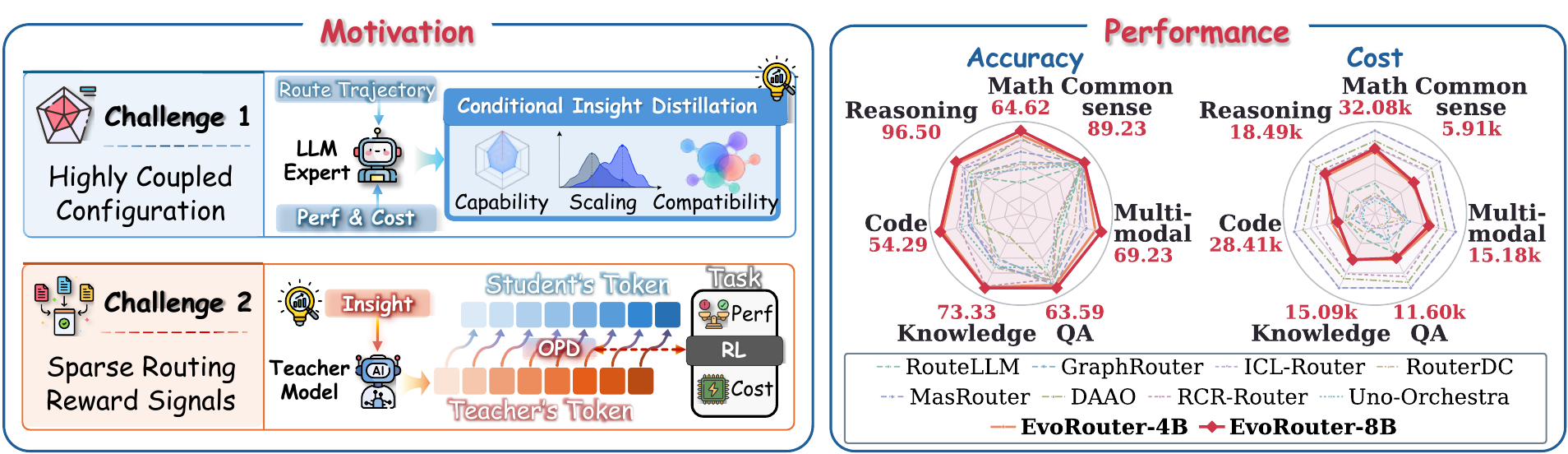}
  \caption{Motivation and performance of EvoRouter. Left: Two challenges and our solutions. Right: EvoRouter's performance.}
  \label{fig:motivation}
\end{figure*}

Agents powered by large language models (LLMs) are rapidly expanding the capabilities of Web services \cite{moslem2026dynamic,li2026deepagent}, from information seeking to complex reasoning and problem solving~\cite{pesl2025advanced,izzi2025data}. LLM routers have emerged as a key component of agentic infrastructure, selecting suitable models for individual queries to balance response quality and inference cost \cite{chen2024routerdc, feng2025graphrouter, wang2026iclrouter, ong2025routellm}. With the growing adoption of multi-agent systems (MAS) in agentic web services such as Grok Bot and WorkBuddy, routing must extend beyond model selection to jointly determine model assignments, collaboration modes, agent counts, interaction rounds, and the skills and subtasks assigned to each agent. For service providers handling Web queries across diverse domains and difficulty levels, adaptively selecting MAS configurations to jointly optimize reasoning performance and compute cost has become a central research problem.%

Existing MAS routing research primarily improves routing decisions through two mechanisms. Parametric policy optimization internalizes execution feedback into router parameters through reinforcement learning or feedback-conditioned distillation, learning policies for model assignment and system orchestration \cite{yue2025masrouter, su2025daao, ke2026masorchestra, feng2026orchmas}. Explicit experience evolution updates retrievable execution records, routing rules, or orchestration skills, allowing accumulated experience to guide subsequent decisions without gradient updates to the router \cite{zhang2026evoroute, rajib2026learning, lin2026skill, li2026rsi}. Connecting experience reuse with policy learning is particularly challenging for joint MAS configuration: the benefits of a model, collaboration mode, or resource allocation depend jointly on task requirements and the surrounding configuration choices. Reusable experience must therefore preserve these conditions of applicability, while effective policy learning must translate the relevant knowledge into the ability to plan new configurations. As illustrated in Figure~\ref{fig:challenge}, this raises a central open question: How can configuration experience be distilled into reusable knowledge of conditional dependencies and internalized in the router’s pre-execution planning policy to improve the system-level trade-off between accuracy and cost?

Addressing this question presents two challenges, as highlighted in Figure~\ref{fig:motivation}. \textbf{(i) Discovering reusable routing principles within a complex, multidimensional configuration space.} Jointly selecting models, collaboration modes, agent counts, interaction rounds, and skills causes a combinatorial explosion in the configuration space, making exhaustive evaluation computationally prohibitive \cite{wang2025agenttts}. Moreover, because configuration benefits depend on both the task and the combination of choices, a single execution trajectory reveals only the outcome under a specific configuration \cite{yu2026recal}. It provides insufficient evidence to identify necessary, redundant, or suboptimal decisions. The router must therefore move beyond learning from isolated outcomes and analyze execution trajectories across tasks and configurations to uncover conditional configuration patterns that generalize to subsequent tasks. \textbf{(ii) Translating post-execution experience into fine-grained supervision for pre-execution routing.} Existing reinforcement learning approaches to routing derive reward signals solely from highly sparse feedback on final task success and computational cost \cite{zhang2025routerr1, cui2026unoorchestra, liaoalpharouter}. However, a complete MAS configuration involves multiple coupled planning decisions. Terminal rewards provide limited guidance on which decisions are effective, which waste resources, and how individual configuration choices should be adjusted. Even when valuable routing insights can be distilled from historical trajectories, they do not automatically translate into effective action selection at specific decision states of the current routing policy. This mismatch between high-dimensional combinatorial decisions and coarse, sparse execution feedback limits the router’s ability to learn efficiently and improve continually from limited execution experience.

To address these challenges, we propose \textbf{EvoRouter}, an adaptive multi-agent routing framework built around an insight-guided self-improvement loop. \textbf{To address the first challenge}, we conduct preliminary experiments that systematically examine how configuration changes affect reasoning performance and computational cost across tasks, revealing task dependence, diminishing marginal returns, and configuration interdependence. We then design an LLM-based expert that distills routing insights by periodically aggregating configurations, execution outcomes, and costs across tasks. The expert derives conditional insights into task-specific capability bottlenecks, marginal returns from resource scaling, and interactions among configuration choices. These insights guide exploration according to task characteristics, steering the router toward optimal configurations. \textbf{To address the second challenge}, we develop a router training method that combines privileged on-policy distillation (OPD) with cost-aware reinforcement learning (RL). OPD transforms configuration insights accumulated across trajectories into dense, token-level rewards informed by routing insights, while RL optimizes the performance–cost trade-off using end-to-end execution outcomes. These complementary training signals address the challenge of learning interdependent multi-agent configuration decisions from sparse execution feedback.

\definecolor{RWHeader}{HTML}{E5E9ED}
\definecolor{RWmodel}{HTML}{E8EFF7}
\definecolor{RWconfiguration}{HTML}{E9F3EA}
\definecolor{RWexperience}{HTML}{FFF1DC}
\definecolor{RWours}{HTML}{D9EEF0}

\begin{table*}[t!]
\centering
\small
\caption{Representative learning methods for MAS routing.}
\label{tab:related_work}
\setlength{\tabcolsep}{6pt} %
\setlength{\extrarowheight}{1pt} %
\renewcommand{\arraystretch}{1.05}
\renewcommand{\tabularxcolumn}[1]{>{\raggedright\arraybackslash}m{#1}}
\begin{tabularx}{\textwidth}{@{}
    >{\hsize=0.72\hsize}X
    >{\hsize=0.84\hsize}X
    >{\hsize=1.40\hsize}X
    >{\hsize=1.04\hsize}X
@{}}
\toprule
\rowcolor{RWHeader}
\textbf{Method} & \textbf{Routing scope} & \textbf{Learning signal} & \textbf{Training mechanism} \\
\midrule
\rowcolor{RWmodel}
RouteLLM~\cite{ong2025routellm} & Single-model choice & Pairwise preferences & Supervised learning \\
\rowcolor{RWmodel}
Router-R1~\cite{zhang2025routerr1} & Multi-round calls & Format, outcome, cost & Outcome-based RL \\
\rowcolor{RWconfiguration}
MasRouter~\cite{yue2025masrouter} & Modes, roles, models & Utility--cost reward & Policy gradients \\
\rowcolor{RWconfiguration}
DAAO~\cite{su2025daao} & Depth, operators, models & Success, utility, cost & Difficulty calibration \\
\rowcolor{RWconfiguration}
MAS-Orchestra~\cite{ke2026masorchestra} & Holistic MAS design & Task-level rewards & Function-calling GRPO \\
\rowcolor{RWexperience}
Skill-MAS~\cite{lin2026skill} & Meta skill & Rollout contrasts & Nonparametric reflection \\
\rowcolor{RWexperience}
RSI-Router~\cite{li2026rsi} & Subtask models, skills & Historical quality--cost traces & Skill update + Pareto selection \\
\rowcolor{RWours}
\textbf{EvoRouter (ours)} & Joint MAS configuration & Token-level supervision from insights,\newline execution feedback, compute overhead & SFT + RL + \textbf{Insight-guided OPD} \\
\bottomrule
\end{tabularx}
\vspace{-4pt}
\end{table*}

\begin{figure*}[t]
    \centering
    \includegraphics[width=\linewidth]{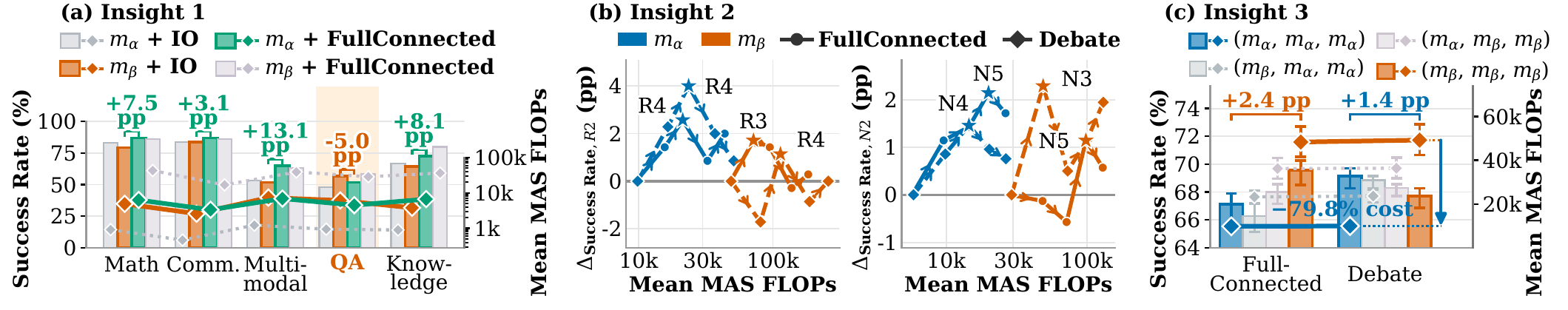}
    \caption{Empirical insights into MAS configuration. (a) Cross-domain sensitivity to different resource scaling; (b) Non-monotonic returns across interaction rounds (left) and agent counts (right). (c) Combinatorial synergy between model compositions and collaboration modes.}
    \label{fig:preexps}
\end{figure*}

Our main contributions are summarized as follows:

\textbf{1. Conditional routing insights distillation.} We identify task dependence, diminishing marginal returns, and interdependence among MAS configuration choices. We distill cross-trajectory performance–cost feedback into configuration insights with explicit applicability conditions to guide exploration.

\textbf{2. Insight-guided router post-training.} We propose EvoRouter, which internalizes conditional insights into the routing policy through OPD and jointly optimizes execution utility with RL, providing fine-grained supervision for routing decisions.

\textbf{3. Comprehensive empirical validation.} We construct a benchmark covering 21 scenarios across seven domains, demonstrating that EvoRouter achieves better performance, lower cost, and greater robustness than representative baselines.

\section{Related Work}
\paragraph{LLM Routing and Multi-Agent Orchestration.} LLM routing spans per-query model selection~\cite{chen2024routerdc,feng2025graphrouter,ong2025routellm,wang2026iclrouter,farahani2026drlm,farahani2026lmedge} and output-budget allocation \cite{xue2026r2}, sequential model invocation \cite{zhang2025routerr1, qian2025xrouter}, and joint MAS configuration \cite{yue2025masrouter, zhao2026sc, su2025daao, ke2026masorchestra, xiao2026routerhgc}. Beyond team design, routers adapt delegation \cite{feng2026orchmas, cui2026unoorchestra}, agent or context selection \cite{zhang2026agentrouter, liu2025rcrrouter}, and execution state \cite{li2026progrouter, rajib2026learning}. Experience-based approaches retain reusable routing knowledge \cite{wang2026skillorchestra, zhang2026evoroute, lin2026skill} or co-evolve routing with worker capabilities \cite{huang2026evolverouter}. RSI-Router\cite{li2026rsi} evolves subtask assignments and model-specific skills through Pareto selection. EvoRouter targets pre-execution joint configuration over fixed heterogeneous workers, informed by our analysis of task-dependent scaling and configuration interactions.

\paragraph{Post-Training for LLMs.} LLM post-training commonly combines supervised fine-tuning (SFT) on demonstrations with preference optimization \cite{ouyang2022training, rafailov2023direct}. Reinforcement learning with verifiable rewards further develops reasoning capabilities from outcome feedback \cite{guo2025deepseek}. Complementing reward-based optimization, on-policy distillation provides teacher supervision on student-generated sequences and can reduce the training–inference distribution mismatch associated with offline distillation \cite{agarwal2024policy}. Recent methods enrich this supervision with textual feedback \cite{hubotter2026reinforcement}, trajectory-derived skills \cite{wang2026skill, wu2026seed}, and role-specific coordination feedback \cite{zhong2026mas}. Building on these principles, EvoRouter combines group-relative RL with OPD on student-generated configuration plans, conditioning the teacher on evolving cross-task insights and current execution feedback. The resulting policy operates without privileged feedback at inference.

\section{Routing Insights and Problem Definition}
\subsection{Routing Insights In Multi-Class Tasks}

\begin{figure*}[t]
  \centering
  \includegraphics[width=\linewidth]{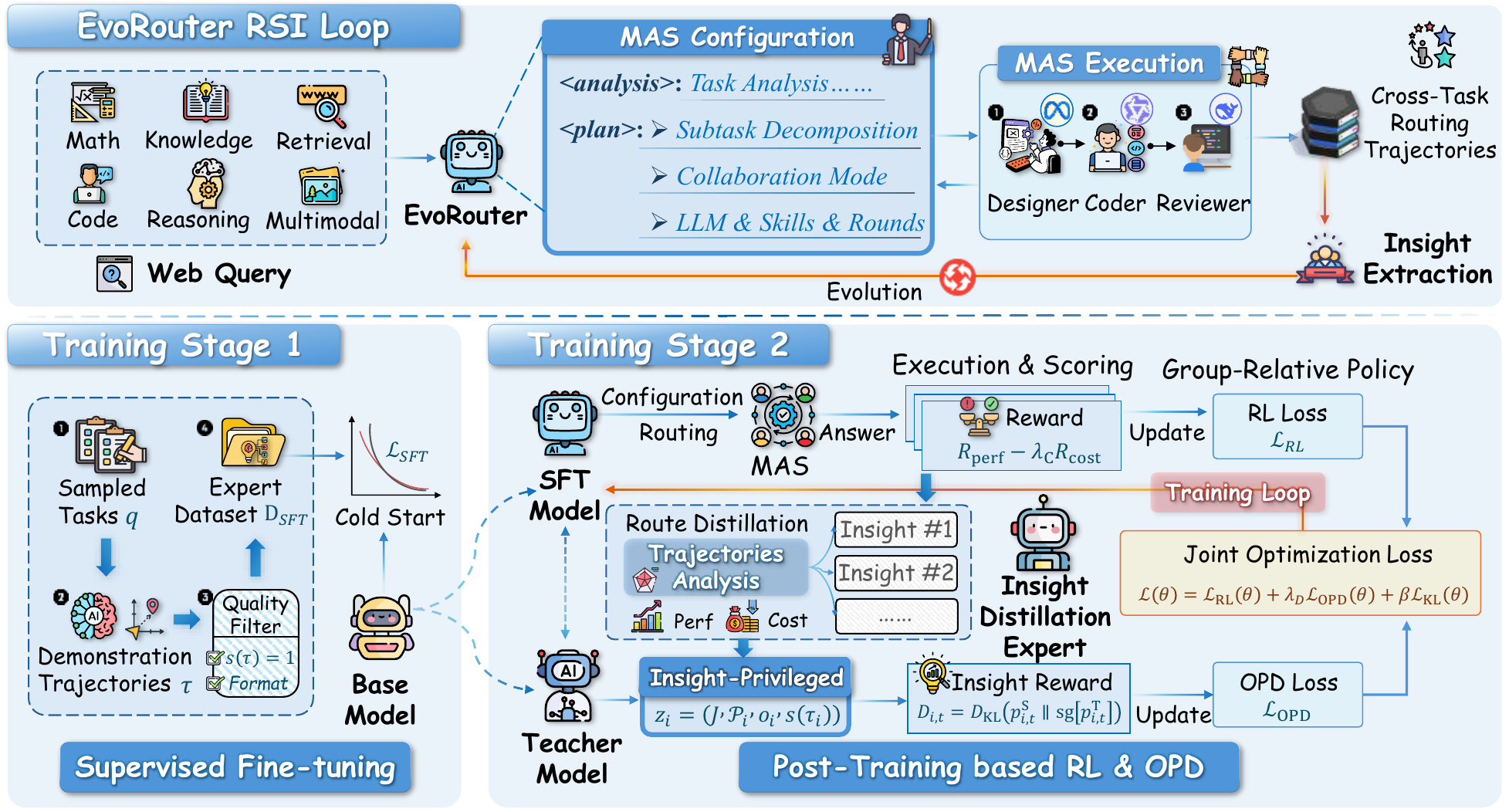}
  \caption{EvoRouter's initialization and evolution loop.}
  \label{fig:overview}
\end{figure*}

\label{sec:pre-experiments}

Efficient MAS routing requires determining \textit{which} resources to scale, \textit{to what extent}, and \textit{how} to combine them. Using $m_\alpha$ (Qwen3.5-35B-A3B) and $m_\beta$ (Qwen3.5-397B-A17B) as representative backbones, we systematically examine how resource selection, scaling scale, and configuration combinations govern task success and compute cost. Figure~\ref{fig:preexps} highlights three key insights. 

\textbf{Where to Scale?} Different tasks exhibit distinct capability demands, leading to task-dependent benefits from resource scaling. As shown in Figure~\ref{fig:preexps}(a), switching from $m_\beta+\mathrm{IO}$ to $m_\alpha+\mathrm{FullConnected}$ improves the success rate on Multimodal by 13.1 points, while reducing that on QA tasks by 5.0 points. This suggests that \textit{scaling decisions should be driven by task demands, prioritizing task-aligned resource configurations} (Insight 1).

\textbf{How Much to Scale?} Increasing resource allocation does not necessarily yield proportional gains. Figure~\ref{fig:preexps}(b) shows that different models and collaboration strategies reach peak performance at different interaction rounds or agent counts, beyond which further scaling may increase costs while degrading success rates. Thus, \textit{resource scaling exhibits configuration-dependent saturation points, and the allocation scale should be tailored to the underlying model and collaboration strategy} (Insight 2).

\textbf{What to Combine?} The effectiveness of individual resources depends on their interactions within the overall configuration. Figure~\ref{fig:preexps}(c) reveals that the relative performance of model compositions can reverse across collaboration strategies: an all-$m_\beta$ configuration outperforms an all-$m_\alpha$ configuration under FullConnected, whereas the latter achieves higher success rates under Debate while reducing MAS FLOPs by 79.8\%. Therefore, \textit{routing insights should capture effective resource combinations and the conditions under which they excel} (Insight 3).

Together, these insights guide subsequent insights distillation from three complementary perspectives: task alignment, scaling saturation, and configuration interactions. Building on these findings, we jointly analyze tasks, complete configurations, execution outcomes, and costs to distill accumulated execution insights into condition-aware routing knowledge.

\subsection{Problem Definition}
\label{sec:problem}
We consider cost-aware multi-agent routing, where a router maps a given task $q\sim\mathcal{Q}$ to a tailored MAS configuration $\mathcal{P}$ drawn from a broad design space spanning models, skills, subtasks, collaboration modes, and agent counts. The router produces a final answer based on the resulting system execution.

Let $\mathcal{M}$, $\mathcal{C}$, and $\mathcal{S}$ denote the available pools of models, collaboration modes, and skills, respectively. Given these resources, the router constructs a detailed MAS configuration $\mathcal{P}$:
\begin{equation}
    \mathcal{P}=\left(c,r,\left\{(u_j,m_j,s_j,p_j)\right\}_{j=1}^{n}\right),
    \label{eq:orchestration-plan}
\end{equation}
where $c\in\mathcal{C}$ specifies the collaboration mode, $r$ is the number of interaction rounds within the system, and $n$ is the number of agents. Each agent $j$ is defined by its identifier $u_j$, model $m_j\in\mathcal{M}$, skill $s_j\in\mathcal{S}$, and assigned subtask $p_j$.

The environment instantiates and executes the MAS, returning an execution outcome $o\sim\mathcal{E}(\cdot\mid q,\mathcal{P})$ to the router, where $\mathcal{E}(\cdot\mid q,\mathcal{P})$ denotes the distribution of execution outcomes conditioned on the task and configuration. The router then generates a final answer $\hat{y}$ for task $q$, forming a complete trajectory $\tau$. A task-specific verifier (e.g., sandbox tests for code generation or exact-match evaluation for question answering) evaluates the final answer, with $s(\tau)\in\{0,1\}$ indicating whether it passes verification.

Let $C(\tau)\geq 0$ denote the execution cost of the routed MAS along trajectory $\tau$, and let $B(q)>0$ be a reference budget for task $q$. We aim to learn a routing policy that jointly optimizes task success and resource efficiency:
\begin{equation}
    \pi^\star\in\arg\max_{\pi}
    \mathbb{E}_{\substack{q\sim\mathcal{Q}\\\tau\sim p_{\pi,\mathcal{E}}(\cdot\mid q,\mathcal{C},\mathcal{S},\mathcal{M})}}
    \left[U\left(s(\tau),\frac{C(\tau)}{B(q)};\lambda_C\right)\right],
    \label{eq:orchestration-objective}
\end{equation}
where $p_{\pi,\mathcal{E}}$ denotes the trajectory distribution jointly induced by the routing policy and multi-agent execution, accounting for stochasticity in both policy generation and system execution. The utility function $U$ balances task success against budget-normalized execution cost, with $\lambda_C$ controlling the cost penalty. Its specific form is defined in Section~\ref{sec:method:rl}.

\section{Method}
\label{sec:method}
We parameterize the routing policy as $\pi_\theta$ and initialize it through SFT. We then collect online trajectories across diverse task categories and employ an insight distillation expert to iteratively extract routing insights, which are internalized into the policy through training. This establishes an iterative evolution loop, sustained by reinforcement learning and on-policy distillation in an online environment, as illustrated in Fig.~\ref{fig:overview}.

\subsection{Evolution Loop}
\label{sec:method:runtime}
\paragraph{Routing and execution.} We construct the router context $x$, comprising role specifications, task descriptions and available routing resources. The router first analyzes the task and reasons within the \AnalysisTag{} tag, then generates a structured plan as defined in Eq.~\eqref{eq:orchestration-plan} within the \PlanTag{} tag. A deterministic executor in the environment parses this plan and instantiates the specified MAS.

The MAS then executes the plan in the environment, producing execution outcomes that are returned as observations for the router's subsequent response. The router analyzes these observations within the \AnalysisTag{} tag and generates the final answer $\hat{y}$ within the \AnswerTag{} tag, forming a two-stage trajectory $\tau$. Each trajectory records the task description, multi-agent configuration, execution outcomes, and verifier-evaluated final answer.
\paragraph{Sampling and evolution.} We deploy the router in an online environment and sample execution trajectories from a training set spanning diverse capability dimensions. Building on the observations from our preliminary experiments (Section~\ref{sec:pre-experiments}), we develop an insight distillation expert to analyze these trajectories and continually refine routing insights. The expert integrates newly observed routing patterns from recent online trajectories with accumulated insights, capturing task-dependent preferences for resources and subtask decomposition strategies across different domains, difficulty levels, and budgets. This process enables routing knowledge to evolve continuously as new execution insights accumulate, as detailed in Section~\ref{sec:method:evolution}.

\subsection{Supervised Initialization}
\label{sec:method:sft}
We initialize the routing policy through SFT to establish protocols for task analysis, system planning, and result synthesis. Specifically, a frozen powerful LLM acts as a bootstrap router to collect trajectories across training tasks $\mathcal{Q}_{\mathrm{sft}}$ under context $x$. We retain only demonstrations with valid formats and correct answers to construct $\mathcal{D}_{\mathrm{SFT}}$. Letting $\mathcal{T}(y)$ index the router-generated tokens in sequence $y$, the initialization objective minimizes
\begin{equation}
    \mathcal{L}_{\mathrm{SFT}}(\theta)=-\sum_{(x,y)\in\mathcal{D}_{\mathrm{SFT}}}\sum_{t\in\mathcal{T}(y)}\log\pi_\theta(y_t\mid x,y_{<t}).
    \label{eq:sft}
\end{equation}
Prompt contexts and environment feedback are masked during loss computation. The supervised policy initializes subsequent iterative evolution and provides a frozen reference policy $\pi_{\mathrm{ref}}$.

\subsection{Cost-Aware Policy Optimization}
\label{sec:method:rl}

\paragraph{Reward curation.} For each trajectory $\tau$, the task verifier assigns an outcome reward $R_{\mathrm{perf}}=s(\tau)\in\{0,1\}$. We measure execution compute $C(\tau)$ and impose a cost penalty $R_{\mathrm{cost}}=\log_2(1+C(\tau)/B(q))$, calibrated against a task-specific reference budget $B(q)$. We then instantiate the utility in Eq.~\eqref{eq:orchestration-objective} as the following reward function $R(\tau)$:
\begin{equation}
    R(\tau)=R_{\mathrm{perf}}-\lambda_C R_{\mathrm{cost}},
    \label{eq:reward}
\end{equation}
where $\lambda_C$ controls cost sensitivity. To strictly separate the reward ranges of successful and failed rollouts, we further cap the weighted cost penalty at $c_{\mathrm{max}}\in[0,1)$ to ensure that successful trajectories receive rewards of at least $1-c_{\mathrm{max}}>0$, while unsuccessful trajectories receive nonpositive rewards. Details of cost weighting are provided in Appendix~\ref{app:dat-details}.

\paragraph{Group-relative policy updates.} For each task $q$, we sample $G$ rollouts $\tau_{q,1},\dots,\tau_{q,G}$ and compute normalized group advantages $A_{q,i}$. Let $\mathcal{T}_{q,i}$ index the router-generated tokens in trajectory $\tau_{q,i}$, conditioned on the preceding interaction history $H_{q,i,t}$. Sharing the same token masking as SFT, the policy loss is defined as:
\begin{equation}
    \mathcal{L}_{\mathrm{RL}}(\theta)=-\frac{1}{Z_{\mathrm{RL}}}\sum_q\sum_{i=1}^{G}A_{q,i}\sum_{t\in\mathcal{T}_{q,i}}\log\pi_\theta\left(y_{q,i,t}\mid H_{q,i,t}\right),
    \label{eq:rl}
\end{equation}
where $Z_{\mathrm{RL}}$ is the total number of masked response tokens from trajectories with nonzero advantages in the batch. To prevent policy drift, we regularize $\pi_\theta$ against the frozen reference $\pi_{\mathrm{ref}}$ via a token-level KL penalty:
\begin{equation}
    \mathcal{L}_{\mathrm{KL}}(\theta)=\frac{1}{Z_{\mathrm{KL}}}\sum_q\sum_{i=1}^{G}\sum_{t\in\mathcal{T}_{q,i}}\left[r_{q,i,t}(\theta)-\log r_{q,i,t}(\theta)-1\right],
    \label{eq:kl}
\end{equation}
where $r_{q,i,t}(\theta)=\frac{\pi_{\mathrm{ref}}(y_{q,i,t}\mid H_{q,i,t})}{\pi_\theta(y_{q,i,t}\mid H_{q,i,t})}$ denotes the reference-to-policy probability ratio, and $Z_{\mathrm{KL}}=\sum_{q,i}|\mathcal{T}_{q,i}|$ counts all router response tokens across the batch, including zero-advantage trajectories.

\begin{table*}[t]
  \centering
  \caption{Overall accuracy (\%) on in-domain and out-of-domain benchmarks. The two Average columns report sample-weighted accuracy. Best and second-best scores in each column are shown in \textbf{bold} and \underline{underline}, respectively.}
  \label{tab:overall_results}
  \begingroup
  \small
  \setlength{\tabcolsep}{1.5pt}
  \renewcommand{\arraystretch}{0.97}
  \providecommand{\sdnote}[1]{%
  \hspace{0.05em}{%
    \color{black!55}%
    \fontsize{4.7}{5.1}\selectfont
    \ensuremath{\pm #1}%
  }%
}
\definecolor{EvoVanillaHead}{HTML}{DFE8F3}
\definecolor{EvoVanillaBody}{HTML}{F3F6FB}
\definecolor{EvoSingleHead}{HTML}{D8EFEC}
\definecolor{EvoSingleBody}{HTML}{F0F9F7}
\definecolor{EvoMultiTurnHead}{HTML}{F9E6CE}
\definecolor{EvoMultiTurnBody}{HTML}{FDF6EC}
\definecolor{EvoMultiAgentHead}{HTML}{E7E0F4}
\definecolor{EvoMultiAgentBody}{HTML}{F6F2FC}
\definecolor{EvoOursHead}{HTML}{DCEBD7}
\definecolor{EvoOursBody}{HTML}{EFF7ED}

  \begin{tabularx}{\textwidth}{@{}>{\raggedright\arraybackslash}p{112pt}*{9}{>{\centering\arraybackslash}X}@{}}
    \toprule
    & \multicolumn{6}{c}{\textbf{In-Domain}} & \multicolumn{3}{c}{\textbf{Out-of-Domain}} \\
    \cmidrule(lr){2-7}\cmidrule(lr){8-10}
    \textbf{Method}
    & {\scriptsize\bfseries Math}
    & {\scriptsize\bfseries Commonsense}
    & {\scriptsize\bfseries Multimodal}
    & {\scriptsize\bfseries QA}
    & {\scriptsize\bfseries Knowledge}
    & {\scriptsize\bfseries Average}
    & {\scriptsize\bfseries Code}
    & {\scriptsize\bfseries Reasoning}
    & {\scriptsize\bfseries Average} \\
    \midrule

    \rowcolor{EvoVanillaHead}
    \multicolumn{10}{@{}l}{\textbf{\textit{Vanilla LLMs}}} \\

    \rowcolor{EvoVanillaBody}
    Qwen3.5-35B-A3B
    & 16.15\sdnote{0.77}
    & 85.00\sdnote{1.02}
    & 29.74\sdnote{0.51}
    & 48.46\sdnote{0.89}
    & 46.67\sdnote{2.86}
    & 46.23\sdnote{0.15}
    & 42.29\sdnote{1.14}
    & 61.50\sdnote{2.78}
    & 52.54\sdnote{0.96} \\

    \rowcolor{EvoVanillaBody}
    Qwen3.5-122B-A10B
    & 25.38\sdnote{1.02}
    & 84.62\sdnote{1.33}
    & 27.69\sdnote{1.54}
    & 56.15\sdnote{0.26}
    & 48.21\sdnote{0.51}
    & 50.23\sdnote{0.71}
    & 49.14\sdnote{1.14}
    & 68.00\sdnote{2.18}
    & 59.20\sdnote{1.22} \\

    \rowcolor{EvoVanillaBody}
    DeepSeek-V4-Flash
    & 45.77\sdnote{2.31}
    & 70.77\sdnote{0.77}
    & 6.67\sdnote{1.36}
    & 47.95\sdnote{0.77}
    & 31.79\sdnote{2.24}
    & 43.46\sdnote{0.43}
    & 40.57\sdnote{0.99}
    & 57.50\sdnote{3.46}
    & 49.60\sdnote{2.31} \\

    \rowcolor{EvoVanillaBody}
    Qwen3.5-397B-A17B
    & 36.15\sdnote{1.02}
    & 88.85\sdnote{0.67}
    & 29.74\sdnote{0.51}
    & 57.95\sdnote{0.92}
    & 55.90\sdnote{3.55}
    & 55.23\sdnote{0.77}
    & 50.86\sdnote{0.57}
    & 75.00\sdnote{1.32}
    & 63.73\sdnote{0.80} \\

    \rowcolor{EvoVanillaBody}
    GLM-5
    & 32.31\sdnote{0.67}
    & 87.31\sdnote{2.14}
    & 21.03\sdnote{1.85}
    & 57.18\sdnote{0.44}
    & 44.62\sdnote{2.24}
    & 50.93\sdnote{1.02}
    & 48.57\sdnote{0.00}
    & 70.50\sdnote{2.18}
    & 60.27\sdnote{1.16} \\
    \midrule

    \rowcolor{EvoSingleHead}
    \multicolumn{10}{@{}l}{\textbf{\textit{Single-Agent Routing}}} \\

    \rowcolor{EvoSingleBody}
    RouteLLM~\cite{ong2025routellm}
    & 23.85\sdnote{3.00}
    & 85.00\sdnote{1.33}
    & 37.44\sdnote{1.36}
    & 59.23\sdnote{1.36}
    & 44.62\sdnote{1.36}
    & 51.85\sdnote{0.54}
    & 25.14\sdnote{1.14}
    & 68.50\sdnote{1.00}
    & 48.27\sdnote{0.92} \\

    \rowcolor{EvoSingleBody}
    GraphRouter~\cite{feng2025graphrouter}
    & 40.00\sdnote{1.68}
    & 87.31\sdnote{2.66}
    & 50.26\sdnote{0.51}
    & 57.44\sdnote{1.03}
    & 51.79\sdnote{1.85}
    & 58.00\sdnote{1.40}
    & 33.71\sdnote{0.57}
    & 73.00\sdnote{2.60}
    & 54.66\sdnote{1.16} \\

    \rowcolor{EvoSingleBody}
    ICL-Router~\cite{wang2026iclrouter}
    & 35.00\sdnote{0.38}
    & 87.69\sdnote{0.67}
    & 50.77\sdnote{1.54}
    & 54.10\sdnote{0.68}
    & 46.15\sdnote{2.35}
    & 55.31\sdnote{0.35}
    & 35.43\sdnote{0.00}
    & 67.50\sdnote{3.61}
    & 52.53\sdnote{1.92} \\

    \rowcolor{EvoSingleBody}
    RouterDC~\cite{chen2024routerdc}
    & 38.46\sdnote{1.39}
    & 86.15\sdnote{1.33}
    & 45.64\sdnote{0.51}
    & 56.92\sdnote{0.68}
    & 53.33\sdnote{3.20}
    & 56.84\sdnote{0.40}
    & 23.43\sdnote{0.99}
    & 73.00\sdnote{2.18}
    & 49.87\sdnote{0.96} \\
    \midrule

    \rowcolor{EvoMultiTurnHead}
    \multicolumn{10}{@{}l}{
      \textbf{\textit{Single-Agent Multi-Turn Routing}}
    } \\

    \rowcolor{EvoMultiTurnBody}
    Router-R1~\cite{zhang2025routerr1}
    & 46.54\sdnote{2.14}
    & 50.38\sdnote{1.39}
    & 11.79\sdnote{0.51}
    & 37.95\sdnote{0.44}
    & 24.10\sdnote{0.00}
    & 36.15\sdnote{0.08}
    & 24.00\sdnote{0.57}
    & 52.00\sdnote{0.87}
    & 38.93\sdnote{0.53} \\

    \rowcolor{EvoMultiTurnBody}
    ReCal~\cite{yu2026recal}
    & 40.38\sdnote{1.76}
    & 80.00\sdnote{2.00}
    & 21.54\sdnote{1.85}
    & 38.72\sdnote{0.92}
    & 37.95\sdnote{2.35}
    & 44.62\sdnote{0.13}
    & 26.85\sdnote{0.57}
    & 58.50\sdnote{3.46}
    & 43.73\sdnote{1.87} \\

    \rowcolor{EvoMultiTurnBody}
    xRouter~\cite{qian2025xrouter}
    & 36.92\sdnote{2.52}
    & 71.15\sdnote{4.44}
    & 38.97\sdnote{0.51}
    & 37.69\sdnote{1.18}
    & 46.67\sdnote{4.89}
    & 45.77\sdnote{1.40}
    & 46.29\sdnote{0.99}
    & 81.00\sdnote{4.27}
    & 64.80\sdnote{2.71} \\
    \midrule

    \rowcolor{EvoMultiAgentHead}
    \multicolumn{10}{@{}l}{\textbf{\textit{Multi-Agent Routing}}} \\

    \rowcolor{EvoMultiAgentBody}
    MAS-Orchestra~\cite{ke2026masorchestra}
    & 46.54\sdnote{2.66}
    & \underline{89.62}\sdnote{1.39}
    & 31.28\sdnote{1.85}
    & 41.03\sdnote{0.44}
    & 58.97\sdnote{2.24}
    & 53.08\sdnote{0.31}
    & 44.57\sdnote{1.14}
    & 86.50\sdnote{2.18}
    & 66.93\sdnote{1.33} \\

    \rowcolor{EvoMultiAgentBody}
    OrchMAS~\cite{feng2026orchmas}
    & 41.15\sdnote{2.34}
    & 85.38\sdnote{0.77}
    & 31.28\sdnote{0.51}
    & 37.44\sdnote{0.77}
    & 31.79\sdnote{3.08}
    & 46.00\sdnote{0.80}
    & 26.28\sdnote{0.57}
    & 68.50\sdnote{4.27}
    & 48.80\sdnote{2.40} \\

    \rowcolor{EvoMultiAgentBody}
    DAAO~\cite{su2025daao}
    & 58.46\sdnote{3.15}
    & 83.08\sdnote{0.67}
    & 32.31\sdnote{0.51}
    & 55.90\sdnote{0.44}
    & 23.08\sdnote{0.51}
    & 53.39\sdnote{0.62}
    & 29.14\sdnote{0.57}
    & 76.50\sdnote{2.18}
    & 54.40\sdnote{0.92} \\

    \rowcolor{EvoMultiAgentBody}
    MasRouter~\cite{yue2025masrouter}
    & 48.08\sdnote{3.90}
    & 88.08\sdnote{2.52}
    & 56.41\sdnote{1.85}
    & 60.77\sdnote{0.92}
    & 50.26\sdnote{1.03}
    & 61.46\sdnote{1.50}
    & 34.86\sdnote{0.57}
    & 84.50\sdnote{2.18}
    & 61.33\sdnote{1.16} \\

    \rowcolor{EvoMultiAgentBody}
    RCR-Router~\cite{liu2025rcrrouter}
    & 57.31\sdnote{2.04}
    & \textbf{90.38}\sdnote{0.77}
    & 29.23\sdnote{0.51}
    & 55.90\sdnote{1.18}
    & 71.79\sdnote{1.85}
    & 61.46\sdnote{0.73}
    & 41.14\sdnote{0.57}
    & 86.50\sdnote{3.97}
    & 65.33\sdnote{2.32} \\

    \rowcolor{EvoMultiAgentBody}
    AgentRouter~\cite{zhang2026agentrouter}
    & 42.31\sdnote{2.14}
    & \textbf{90.38}\sdnote{2.04}
    & 44.10\sdnote{1.03}
    & 58.97\sdnote{1.36}
    & 56.41\sdnote{2.66}
    & 59.31\sdnote{0.28}
    & 28.57\sdnote{0.00}
    & 79.00\sdnote{0.87}
    & 55.47\sdnote{0.46} \\

    \rowcolor{EvoMultiAgentBody}
    Uno-Orchestra~\cite{cui2026unoorchestra}
    & 34.23\sdnote{0.77}
    & 85.00\sdnote{1.02}
    & 34.36\sdnote{0.89}
    & 46.15\sdnote{0.68}
    & 52.82\sdnote{0.00}
    & 50.77\sdnote{0.28}
    & 40.00\sdnote{0.00}
    & 73.00\sdnote{1.32}
    & 57.60\sdnote{0.71} \\
    \midrule

    \rowcolor{EvoOursHead}
    \multicolumn{10}{@{}l}{\textbf{\textit{EvoRouter (Ours)}}} \\

    \rowcolor{EvoOursBody}
    \textbf{EvoRouter-4B}
    & 61.15\sdnote{2.04}
    & 88.46\sdnote{0.67}
    & 64.10\sdnote{2.71}
    & 61.03\sdnote{1.56}
    & 72.31\sdnote{2.35}
    & 68.69\sdnote{0.95}
    & 52.57\sdnote{0.57}
    & 94.00\sdnote{1.50}
    & 74.67\sdnote{0.96} \\

    \rowcolor{EvoOursBody}
    \textbf{EvoRouter-8B}
    & \textbf{64.62}\sdnote{1.02}
    & 89.23\sdnote{0.22}
    & \underline{69.23}\sdnote{0.51}
    & \textbf{63.59}\sdnote{1.07}
    & \underline{73.33}\sdnote{2.05}
    & \textbf{71.23}\sdnote{0.74}
    & \underline{54.29}\sdnote{1.44}
    & \textbf{96.50}\sdnote{1.26}
    & \underline{76.80}\sdnote{0.67} \\

    \rowcolor{EvoOursBody}
    \textbf{EvoRouter-LLaMa-8B}
    & \underline{63.46}\sdnote{1.94}
    & 89.23\sdnote{2.62}
    & \textbf{69.74}\sdnote{1.85}
    & \underline{62.05}\sdnote{1.29}
    & \textbf{76.92}\sdnote{2.07}
    & \underline{71.15}\sdnote{0.89}
    & \textbf{59.43}\sdnote{0.40}
    & \underline{96.00}\sdnote{1.77}
    & \textbf{78.93}\sdnote{1.13} \\

    \bottomrule
  \end{tabularx}
  \endgroup
\end{table*}

\subsection{Insight-Guided On-Policy Distillation}
\label{sec:method:evolution}

\paragraph{Recursive insight consolidation.} Guided by the analytical framework in Section~\ref{sec:pre-experiments}, we initialize an insight pool $\mathcal{I}_0$ and employ an expert $\mathcal{F}$ to periodically distill execution observations into generalizable routing rules with explicit applicability conditions only during training. For the $k$-th insight update using recent rollout batch $\mathcal{B}_k$, the expert refines the insight set through a three-step procedure:
\textit{(i) Rule induction:} comparing intra-category trajectories across success or failure and cost spectra to derive conditional rules on capability alignment, scaling saturation, and configuration interactions;
\textit{(ii) Precondition refinement:} resolving empirical conflicts within $\mathcal{I}_k$ by tightening boundaries, such as specifying finer-grained task difficulty and domains; and
\textit{(iii) Capacity pruning:} merging redundant entries and enforcing a strict capacity ceiling to curb context inflation. Formally, the update is expressed as:
\begin{equation}
    \mathcal{I}_{k+1}=\mathcal{F}\left(\mathcal{I}_k,\mathcal{B}_k\right).
    \label{eq:insight-update}
\end{equation}
Distillation steps always condition on the latest insights $\mathcal{I}$. Further details on insight distillation are provided in Appendix~\ref{app:insights}.

\paragraph{Insight-conditioned supervision.} For each trajectory $\tau_i$ sampled from the student policy, the frozen teacher model $\pi_\phi$ receives additional privileged information $z_i=(\mathcal{I},\mathcal{P}_i,o_i,s(\tau_i))$, where $o_i$ denotes the MAS execution outcome, while sharing the same context $x_i$ and prefix $y_{i,<t}$ as the student. Specifically, at each token position $t$ in the routing response, let $p_{i,t}^{\mathrm{S}}=\pi_\theta(\cdot\mid x_i,y_{i,<t})$ and $p_{i,t}^{\mathrm{T}}=\pi_\phi(\cdot\mid x_i,z_i,y_{i,<t})$ denote the next-token distributions of the student and teacher, respectively. Following MiniLLM~\cite{minillm}, we minimize the reverse KL divergence over the vocabulary $\Omega$ shared by the teacher and student:
\begin{equation}
    D_{i,t}=D_{\mathrm{KL}}\left(p_{i,t}^{\mathrm{S}}\,\middle\|\,\operatorname{sg}\left[p_{i,t}^{\mathrm{T}}\right]\right)=\sum_{a\in\Omega}p_{i,t}^{\mathrm{S}}(a)\log\frac{p_{i,t}^{\mathrm{S}}(a)}{\operatorname{sg}\left[p_{i,t}^{\mathrm{T}}(a)\right]},
\end{equation}
where $D_{\mathrm{KL}}(·)$ denotes the KL divergence, and $\operatorname{sg}[\cdot]$ denotes the stop-gradient operator.

\paragraph{Insight internalization and joint optimization.} Distillation is applied to the initial routing response of each trajectory. Let $\mathcal{T}_i^{\mathrm{ana}}$ and $\mathcal{T}_i^{\mathrm{plan}}$ denote the positions of student-generated content tokens within the \AnalysisTag{} and \PlanTag{} tags, respectively, excluding formatting and environment tokens. The OPD objective minimizes their weighted distillation losses:
\begin{equation}
    \mathcal{L}_{\mathrm{OPD}}(\theta)
    =\frac{\sum_i\sum_{t\in\mathcal{T}_i^{\mathrm{plan}}}D_{i,t}}{\sum_i|\mathcal{T}_i^{\mathrm{plan}}|}
    +\alpha\frac{\sum_i\sum_{t\in\mathcal{T}_i^{\mathrm{ana}}}D_{i,t}}{\sum_i|\mathcal{T}_i^{\mathrm{ana}}|},
\end{equation}
where $\alpha$ controls the contribution of the analysis tokens.

The final joint training objective is defined as:
\begin{equation}
    \mathcal{L}(\theta)=\mathcal{L}_{\mathrm{RL}}(\theta)+\lambda_D\mathcal{L}_{\mathrm{OPD}}(\theta)+\beta\mathcal{L}_{\mathrm{KL}}(\theta),
\end{equation}
where $\lambda_D$ and $\beta$ are the coefficients for distillation and KL regularization, respectively. The updated policy $\pi_\theta$ continues to generate execution trajectories that inform subsequent insight refinement, closing the recursive feedback loop among execution, insight synthesis, and policy internalization.

\begin{figure*}[t]
  \centering
  \includegraphics[width=\linewidth]{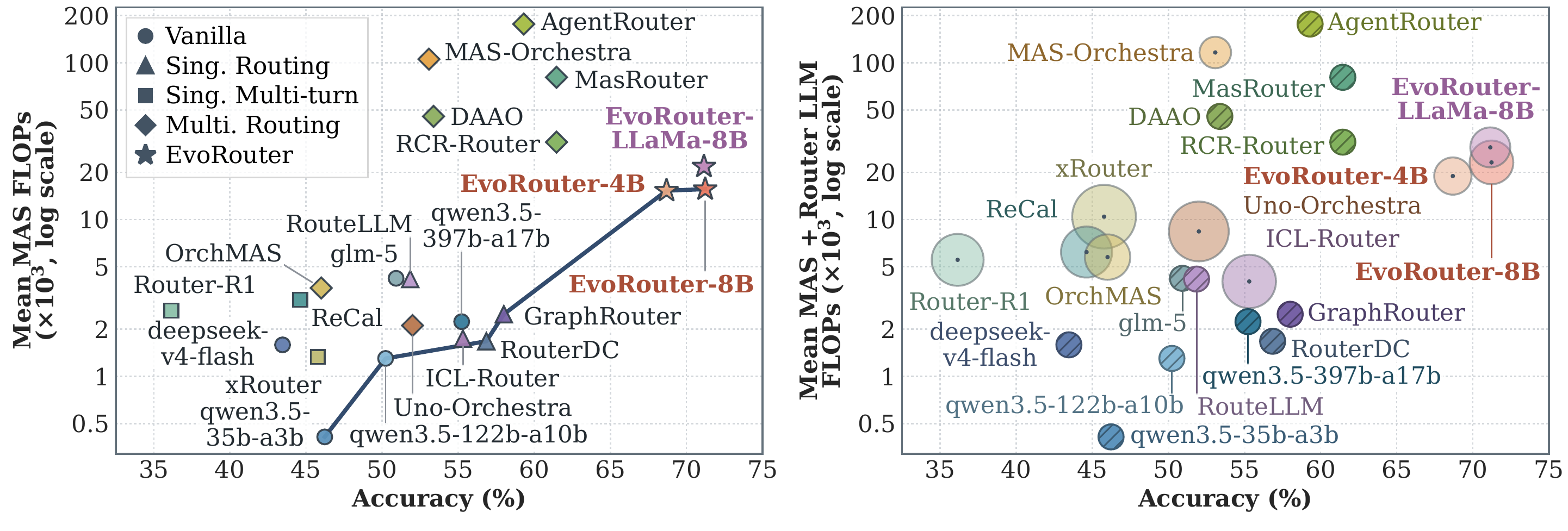}
  \caption{Left: The Pareto frontier of the evaluated baselines; Right: Mean total LLM inference cost, including router calls.}
  \label{fig:cost_tradeoff}
\end{figure*}

\section{Experiments}

\subsection{Experimental Setup}
\label{sec:experimental_setup}

\paragraph{Datasets and metrics.} We evaluate EvoRouter across 21 datasets in seven domains: math (MATH~\cite{hendrycks2021math}, GSM8K~\cite{cobbe2021gsm8k}, Omni-MATH-2~\cite{ballon2026omnimath2}, Apex Shortlist~\cite{dekoninck2026matharena}), commonsense (ARC-Challenge~\cite{clark2018arc}, StrategyQA~\cite{geva2021strategyqa}, Social IQA~\cite{sap2019socialiqa}), knowledge (MMLU-Pro~\cite{wang2024mmlupro}, GPQA-Diamond~\cite{rein2024gpqa}, SuperGPQA~\cite{du2025supergpqa}), retrieval (NQ~\cite{kwiatkowski2019naturalquestions}, TriviaQA~\cite{joshi2017triviaqa}, HotpotQA~\cite{yang2018hotpotqa}, 2WikiMultiHopQA~\cite{ho2020twowikimultihopqa}, PopQA~\cite{mallen2023popqa}, MuSiQue~\cite{trivedi2022musique}), multimodal (ChartQAPro~\cite{masry2025chartqapro}, MMMU~\cite{yue2024mmmu}, MMMU-Pro~\cite{yue2025mmmupro}), alongside two out-of-domain tasks: code generation (LiveCodeBench~\cite{jain2025livecodebench}) and logical reasoning (BBH-LD~\cite{suzgun2023bbh}). Table~\ref{tab:overall_results} reports verifier-based accuracy detailed in Section~\ref{sec:problem}. We measure the per-question computational cost of each MAS using the relative FLOPs metric defined in Appendix~\ref{app:compute-cost}.

\paragraph{Routing resources.} The available LLM pool comprises Qwen3.5-35B-A3B, Qwen3.5-122B-A10B, Qwen3.5-397B-A17B, DeepSeek-V4-Flash, and GLM-5. The collaboration mode pool includes IO, CoT, Reflection, Chain, FullConnected, and Debate. Each task category is additionally equipped with at least six relevant skills, such as code verification for coding tasks and information synthesis for QA. All routing methods share the same resources and are provided with identical descriptions of available LLMs, collaboration modes, and skills wherever applicable.

\paragraph{Baselines.} We compare five categories of methods. \textit{(I) Vanilla LLMs:} Direct inference using individual models from the LLM pool. \textit{(II) Single-Agent Routing:} RouteLLM~\cite{ong2025routellm}, GraphRouter~\cite{feng2025graphrouter}, ICL-Router~\cite{wang2026iclrouter}, and RouterDC~\cite{chen2024routerdc}. \textit{(III) Single-Agent Multi-Turn Routing:} Router-R1~\cite{zhang2025routerr1}, ReCal~\cite{yu2026recal}, and xRouter~\cite{qian2025xrouter}. \textit{(IV) Multi-Agent Routing:} MAS-Orchestra~\cite{ke2026masorchestra}, OrchMAS~\cite{feng2026orchmas}, DAAO~\cite{su2025daao}, MasRouter~\cite{yue2025masrouter}, RCR-Router~\cite{liu2025rcrrouter}, AgentRouter~\cite{zhang2026agentrouter}, and Uno-Orchestra~\cite{cui2026unoorchestra}. \textit{(V) Ours:} EvoRouter-4B, EvoRouter-8B, and EvoRouter-LLaMa-8B. Further details on the baselines are provided in Appendix~\ref{app:eva-details}.

\subsection{Performance Analysis}

\paragraph{Accuracy and cross-domain generalization.}
As shown in Table~\ref{tab:overall_results}, all EvoRouter variants outperform competing baselines across four of the five in-domain capability categories, achieving average accuracies of 68.69\% for EvoRouter-4B and 71.23\% for EvoRouter-8B. \textbf{This advantage extends to unseen Code and Reasoning domains}, where EvoRouter achieves sample-weighted accuracies of 74.67\% to 78.93\%, surpassing the strongest baseline by 7.74 to 12.00 percentage points. These consistent gains highlight the transferability of EvoRouter's learned orchestration strategies across diverse capability domains.

\paragraph{Compute efficiency.}
The left panel of Figure~\ref{fig:cost_tradeoff} highlights EvoRouter's favorable position on the accuracy–cost Pareto frontier. EvoRouter-4B achieves 7.23 percentage points higher accuracy than the strongest baseline RCR-Router while reducing mean MAS cost by 50.95\% to 15.28k FLOPs units. \textbf{Accounting for router overhead,} the right panel shows that total inference cost reaches only 18.94k units, remaining 76.5\% and 39.2\% below MasRouter and RCR-Router, respectively. This confirms that substantial worker-level compute savings translate into end-to-end efficiency gains despite the additional planning overhead.

\begin{table}[t]
    \centering
    \caption{Component ablations of EvoRouter-4B.}
    \label{tab:training_ablation}
    \begingroup
    \small
    \setlength{\tabcolsep}{4pt}
    \providecommand{\sdnote}[1]{%
    \hspace{0.05em}{%
        \color{black!55}%
        \fontsize{4.7}{5.1}\selectfont
        \ensuremath{\pm #1}%
    }%
    }
    \renewcommand{\arraystretch}{1.15}
    \resizebox{\columnwidth}{!}{%
        \begin{tabular}{@{}lcccc@{}}
            \toprule
            \multirow{2}{*}{\textbf{Training Stage}}
                & \multicolumn{2}{c}{\textbf{In-Domain}}
                & \multicolumn{2}{c}{\textbf{Out-of-Domain}} \\
            \cmidrule(lr){2-3} \cmidrule(l){4-5}
                & Accuracy & Cost & Accuracy & Cost \\
            \midrule
            EvoRouter-4B
                & 68.69\sdnote{0.95}
                & 15280.79\sdnote{770.31}
                & 74.67\sdnote{0.96}
                & 25952.47\sdnote{2673.56} \\

            ~~~ \textit{w/o Insights}
                & 64.54\sdnote{1.52}
                & 18245.46\sdnote{977.52}
                & 70.93\sdnote{2.44}
                & 25453.12\sdnote{2075.82} \\

            ~~~ \textit{w/o OPD}
                & 63.23\sdnote{1.31}
                & 16467.21\sdnote{852.09}
                & 69.87\sdnote{1.41}
                & 24419.74\sdnote{2017.43} \\

            ~~~ \textit{w/o RL+OPD}
                & 59.77\sdnote{0.38}
                & 16789.73\sdnote{691.28}
                & 63.20\sdnote{1.16}
                & 30158.45\sdnote{2555.80} \\

            ~~~ \textit{w/o SFT+RL+OPD}
                & 53.23\sdnote{0.23}
                & 25472.25\sdnote{1278.80}
                & 52.80\sdnote{2.40}
                & 19795.09\sdnote{1654.15} \\
            \bottomrule
        \end{tabular}%
    }
    \endgroup
\end{table}

\subsection{Framework Analysis}

\paragraph{Ablation study.}
Table~\ref{tab:training_ablation} quantifies the contributions of RL, OPD, and distilled Insights. Compared with SFT alone, full EvoRouter-4B improves in-domain and out-of-domain accuracy by 8.92 and 11.47 percentage points while reducing MAS execution costs by 9.0\% and 13.9\%, respectively. Under matched training iterations, OPD without Insights yields only 1.31- and 1.06-point accuracy gains over RL, whereas insight-guided OPD boosts these gains to 5.46 and 4.80 points. This substantial gap highlights the importance of distilling cross-trajectory insights into actionable supervision for effective routing.

\begin{figure}[htbp]
    \centering
    \includegraphics[width=\linewidth]{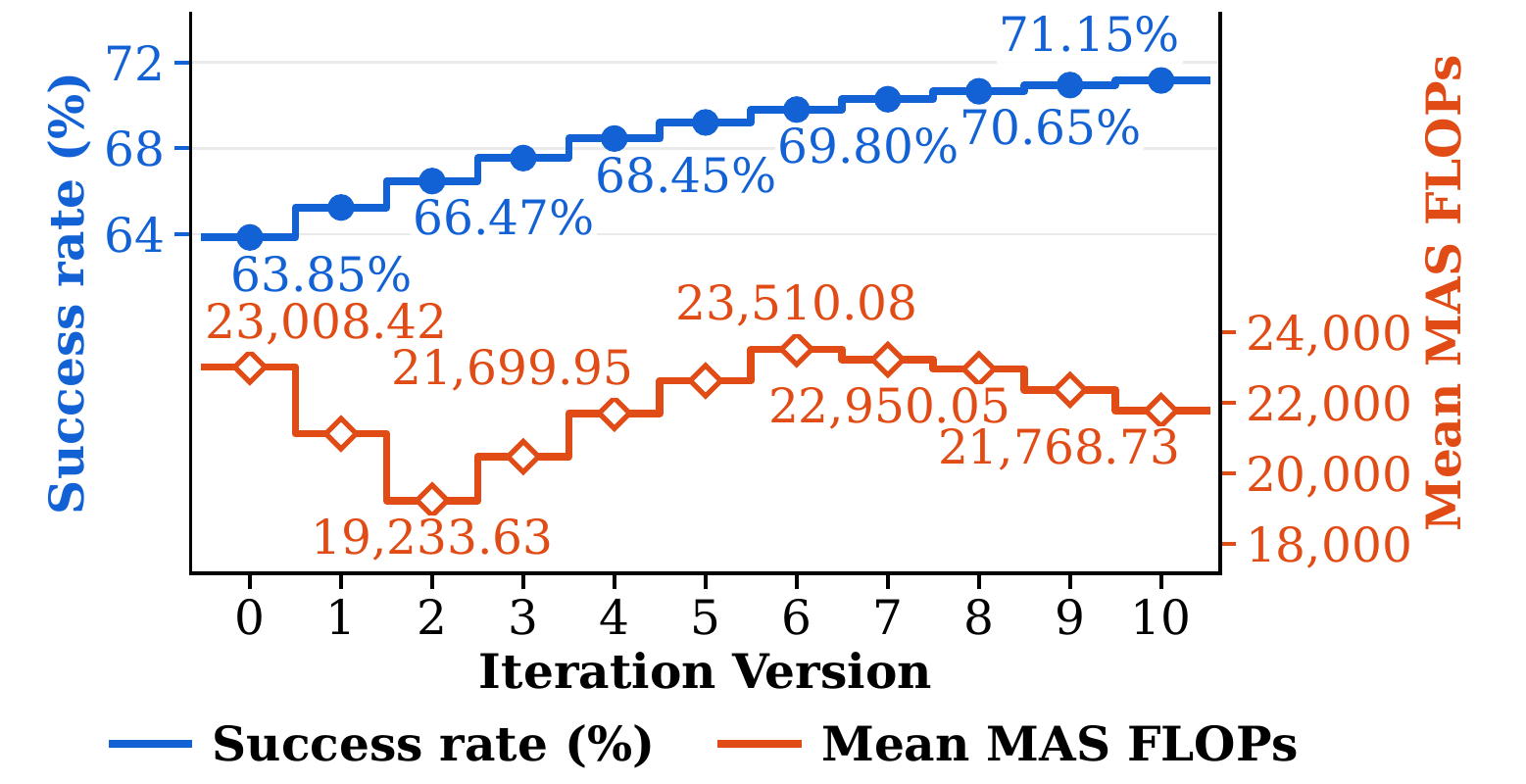}
    \caption{Training dynamics of EvoRouter-LLaMa-8B.}
    \label{fig:vers}
\end{figure}

\paragraph{Evolution dynamics.}
Figure~\ref{fig:vers} illustrates how iterative insight-guided policy refinement progressively improves EvoRouter-LLaMa-8B's performance. Early updates improve success rate by 2.62 points while reducing compute by 16.4\%, reflecting more efficient resource selection; Subsequent updates leverage additional compute for further accuracy gains, before later refinements achieve another 1.35-point improvement with 7.4\% lower cost. These dynamics demonstrate how insight-guided evolution progressively refines the trade-off between accuracy and computational efficiency. Training loss curves for all three backbones are provided in Appendix~\ref{app:tra-details}.

\begin{figure}[htbp]
    \centering
    \includegraphics[width=\linewidth]{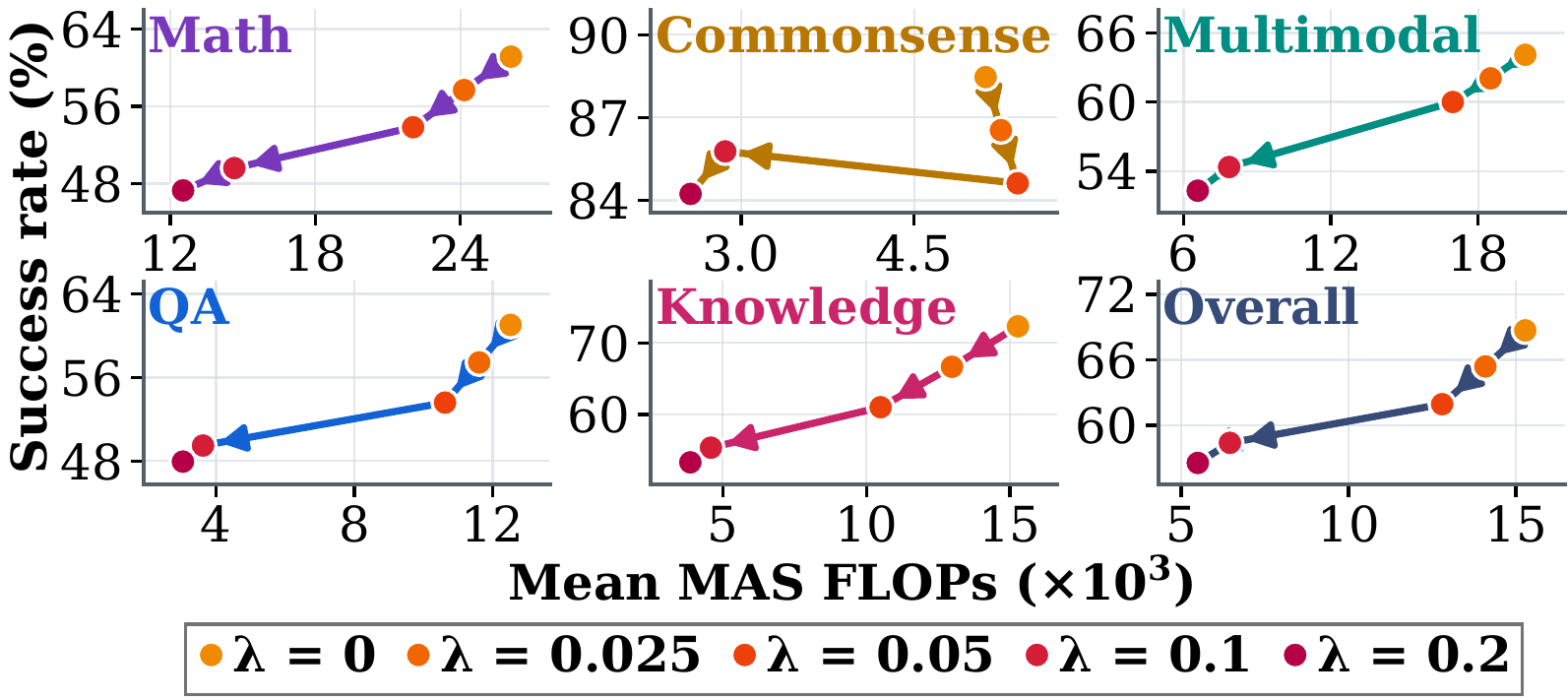}
    \caption{Sensitivity of EvoRouter-4B to coefficient $\lambda_C$.}
    \label{fig:sensivity}
\end{figure}

\begin{figure}[htbp]
    \centering
    \includegraphics[width=\linewidth]{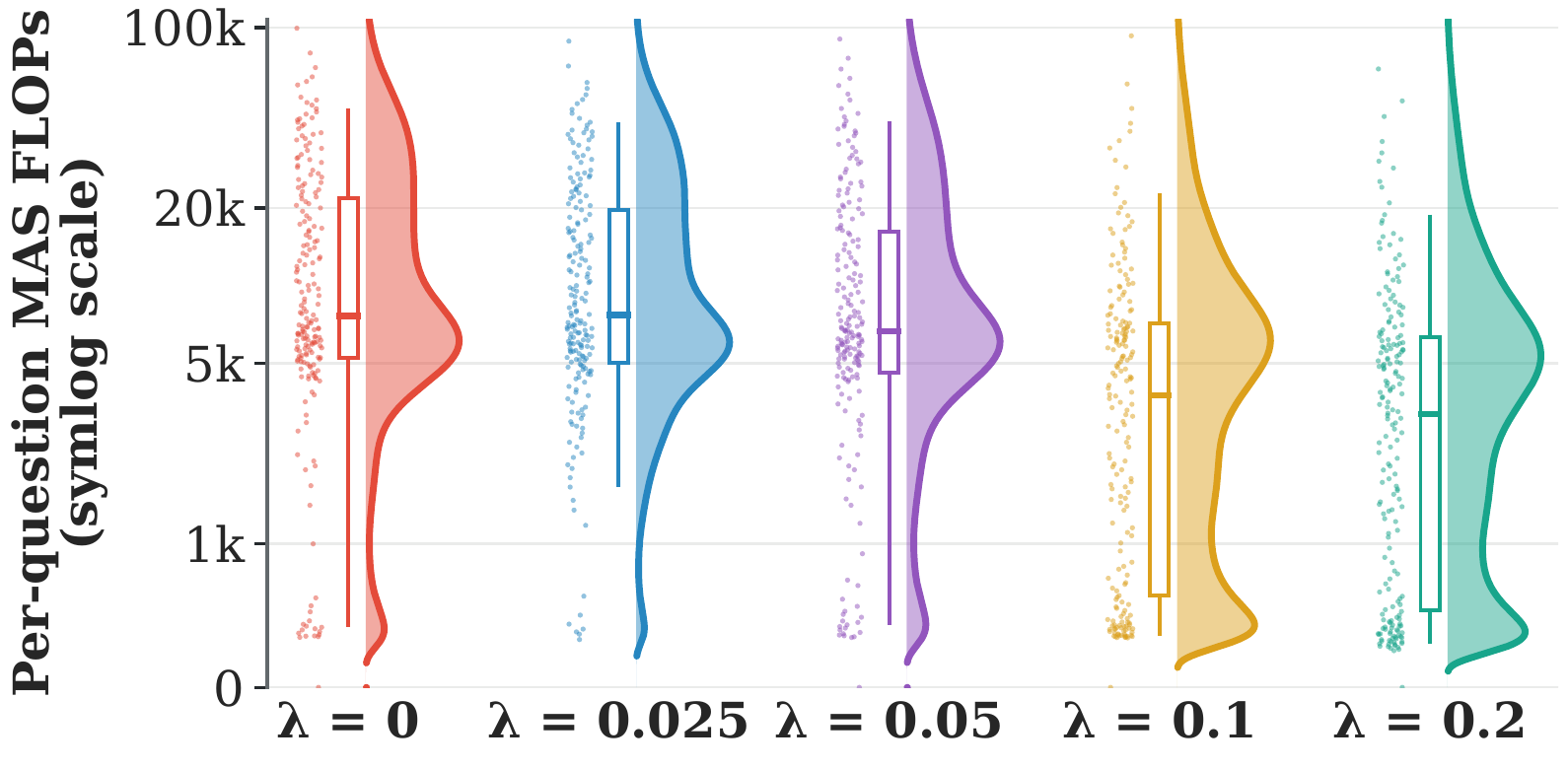}
    \caption{Per-question MAS cost distributions of EvoRouter-4B under different cost penalties.}
    \label{fig:raincloud}
\end{figure}

\paragraph{Sensitivity study.}
We independently train EvoRouter-4B across $\lambda_C\in\{0,0.025,0.05,0.1,0.2\}$. As shown in Figure~\ref{fig:sensivity}, higher $\lambda_C$ steers the policy toward lower-cost configurations, with varying accuracy sensitivity across domains. Furthermore, Figure~\ref{fig:raincloud} reveals that increasing cost penalties progressively shifts the per-question compute distribution toward lower-budget regimes while substantially suppressing the expensive upper tail, ensuring predictable workload expenditure.

\subsection{Generalization Analysis}
\label{sec:resource-generalization}
Beyond the two out-of-domain datasets reported in Table~\ref{tab:overall_results}, we further evaluate zero-shot generalization to unseen routing resources by simultaneously replacing the LLM, collaboration mode, and skill pools while keeping the trained router fixed. Replacement resources are provided in Appendix~\ref{app:res-details}. As shown in Figure~\ref{fig:generality}, EvoRouter-4B consistently achieves the highest accuracy, outperforming MasRouter by 3.85 and 7.47 percentage points on in-domain and out-of-domain tasks while reducing MAS compute by 6.7\% and 18.5\%, respectively. Notably, it also surpasses the strongest out-of-domain baseline by 5.9 points with less than half its compute expenditure. Compared with the independently trained EvoRouter-4B \textit{w/o Insights} variant, the full model achieves 5.77 and 11.47 percentage points higher accuracy, with the larger out-of-domain gain highlighting the role of insight-guided training in transferring orchestration knowledge across unseen resources and task distributions.

\begin{figure}[htbp]
    \centering
    \includegraphics[width=\linewidth]{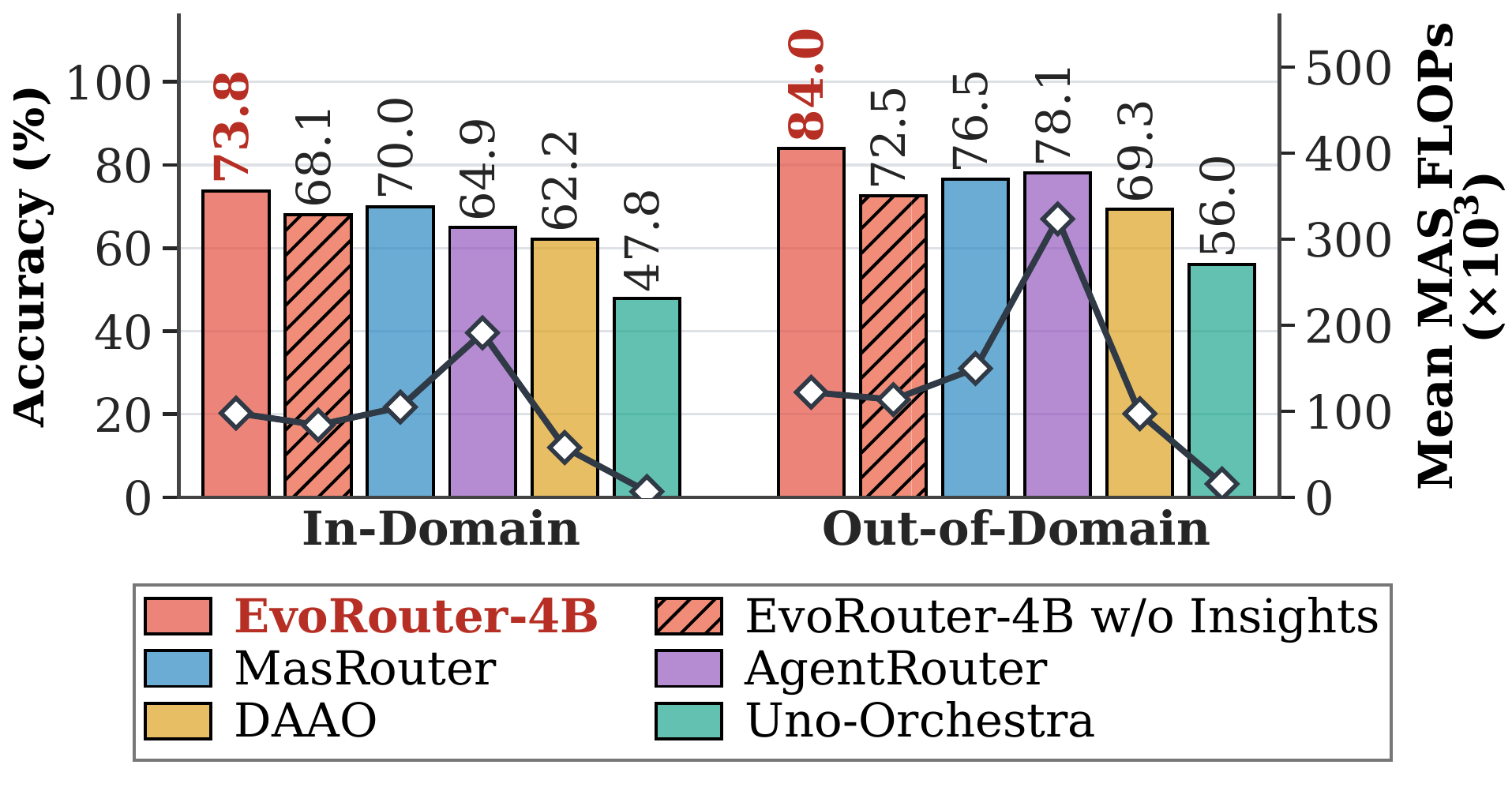}
    \caption{Generalization to unseen routing resources.}
    \label{fig:generality}
\end{figure}

\subsection{Case Study}
\begin{figure}[htbp]
    \centering
    \includegraphics[width=0.98\linewidth]{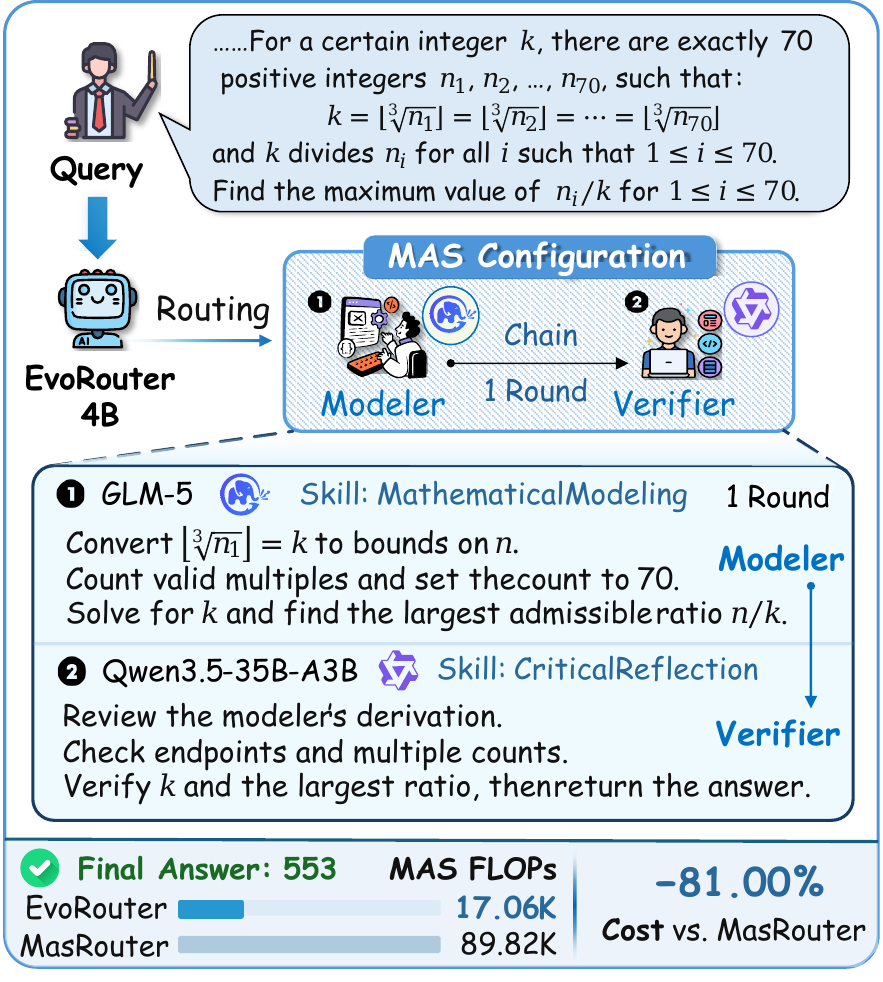}
    \caption{A practical case of EvoRouter.}
    \label{fig:case}
\end{figure}

To better illustrate the routing mechanism of EvoRouter, we present a representative case on mathematical reasoning in Figure~\ref{fig:case}, demonstrating how it achieves accurate and cost-effective MAS routing. An illustrative example of the distilled routing insights is provided in Figure~\ref{fig:evolution-insights}.

\section{Conclusion}

We presented EvoRouter, a framework that learns to jointly configure heterogeneous LLM-based MAS from execution experience. Combining cost-aware reinforcement learning with on-policy distillation, EvoRouter internalizes cross-task configuration insights and trajectory feedback into its routing policy, enabling deployment without privileged feedback. Across 21 datasets spanning seven domains, EvoRouter achieves higher average accuracy at lower computational cost than the evaluated routing baselines in both in-domain and out-of-domain settings. Our current framework is designed for cloud-based agent services. Future work will extend EvoRouter to resource-constrained edge environments through coordinated cloud–edge routing that accounts for device heterogeneity and inter-agent communication overhead.

\begingroup
\raggedright
\bibliographystyle{ACM-Reference-Format}
\bibliography{references}
\endgroup

\appendix

\section{Experimental Details}
\label{app:details}

\subsection{Resource}
\label{app:res-details}
All LLMs in the pool use temperature 1.0 and maximum output length 1,024 tokens (4,096 for code tasks). For multimodal tasks, models lacking multimodal support receive a fixed placeholder declaring image invisibility, forcing them to answer solely based on text. In generalization experiments (Section~\ref{sec:resource-generalization}), the model pool comprises Qwen3.8-Flash, Kimi-K2.8-Preview, and Qwen3.8-Max; the collaboration mode pool comprises Self-Consistency and Hierarchical; and the skill pool contains skills unused elsewhere.

Details of the collaboration modes are as follows:

\begin{itemize}
  \item \textbf{IO}: A single agent answers in a single turn.
  \item \textbf{CoT}: A single agent reasons step-by-step across multiple turns before delivering the final answer.
  \item \textbf{Reflection}: A single agent iteratively reflects on its previous response to generate a revised answer.
  \item \textbf{Self-Consistency}: A single agent is invoked multiple times independently. A subsequent invocation then inspects all candidate answers, evaluates consistency, resolves discrepancies, and aggregates the final result.
  \item \textbf{Chain}: Agents are activated sequentially, each receiving its predecessor's output. When an agent exhausts its allocated reasoning turns, the next agent is triggered.
  \item \textbf{FullConnected}: All agents are activated concurrently, receiving peers' responses from the prior turn. Final-turn responses are routed to the router to synthesize the answer.
  \item \textbf{Debate}: All agents run concurrently, critiquing peers or rebutting critiques in subsequent turns. Final-turn responses are routed to the router for answer synthesis.
  \item \textbf{Hierarchical}: The initial agent plans task decomposition and scheduling, intermediate agents execute independently in parallel, and the final agent synthesizes findings. Turn limits govern reasoning turns per agent.
\end{itemize}

Certain collaboration modes impose cross-configuration constraints. For example, if the router pairs CoT with two agents, the MAS plan is rejected (as CoT admits at most one agent) and marked incorrect.

\subsection{Dataset}
\label{app:dat-details}
We construct a dataset comprising 12,800 questions across five domains, as described in Section~\ref{sec:experimental_setup}. The dataset is partitioned into 10,500 training, 1,000 development, and 1,300 test questions. For out-of-domain evaluation, we use 175 questions from the LiveCodeBench~\cite{jain2025livecodebench} and 200 from the BBH-LD~\cite{suzgun2023bbh}.

For each training task domain, we set a shared budget in our cost units. We calibrate it from pre-collected successful traces by computing the 25th percentile of trace costs for each question, taking the median across questions in the domain, and adding a 15\% margin, with a minimum budget of 512. Budgets vary across domains (as well as trace collections); for example, the budget for the math tasks is 12,374.00, and 5,722.02 for commonsense tasks.

\subsection{Training}
\label{app:insights}
\label{app:tra-details}
We evaluate three router backbones: Qwen3-4B, Qwen3-8B, and Llama-3.1-8B, yielding EvoRouter-4B, EvoRouter-8B, and EvoRouter-LLaMa-8B, respectively, through our iterative training framework. All experiments run on 8 NVIDIA A100 80GB GPUs. Task sets across SFT, RL, and evaluation are mutually disjoint.

For supervised initialization, DeepSeek-V4-Pro serves as a bootstrap router to collect trajectories, yielding 2,000 filtered instances to fine-tune each backbone for 2 epochs. In subsequent training, we use a batch size of 8 with 16 rollouts per question and a learning rate of $3\times10^{-6}$, setting $\beta=0.001$, $c_{\mathrm{max}}=0.1$, $\lambda_D=0.03$, and $\alpha=0.25$. For on-policy distillation, Qwen3-32B and Llama-3.1-70B serve as teachers for the Qwen and Llama backbones, respectively. DeepSeek-V4-Pro also acts as the insight distillation expert, leveraging its context window and reasoning capabilities to analyze dozens of trajectories simultaneously, but does not directly supervise the student. Finally, we cap insight capacity at $|\mathcal{I}| \le 8$ and total length at 6,144 characters to limit context overhead.

\begin{figure}[htbp]
\centering
\begin{tcolorbox}[title={Example Insights from EvoRouter's Evolution Loop}, colback=gray!5, colbacktitle=gray!100, coltitle=white, colframe=gray!60, boxrule=0.4pt, arc=1mm]

\textcolor{analysisfg}{\textit{{Task dependency.}}} Missing or ambiguous facts need models with retrieval or stronger knowledge; $\ldots$

\textcolor{analysisfg}{\textit{{Diminishing marginal returns.}}} For simple arithmetic, accuracy saturates with one capable worker and one round; $\ldots$

\textcolor{analysisfg}{\textit{{Combinatorial dependency.}}} Use Chain for chart calculations, pairing a multimodal extraction worker with a numerical solver consuming verified readings; $\ldots$

\end{tcolorbox}
\caption{Example insights distilled during EvoRouter's evolution.}
\label{fig:evolution-insights}
\end{figure}

\begin{figure}[htbp]
    \centering
    \includegraphics[width=\linewidth]{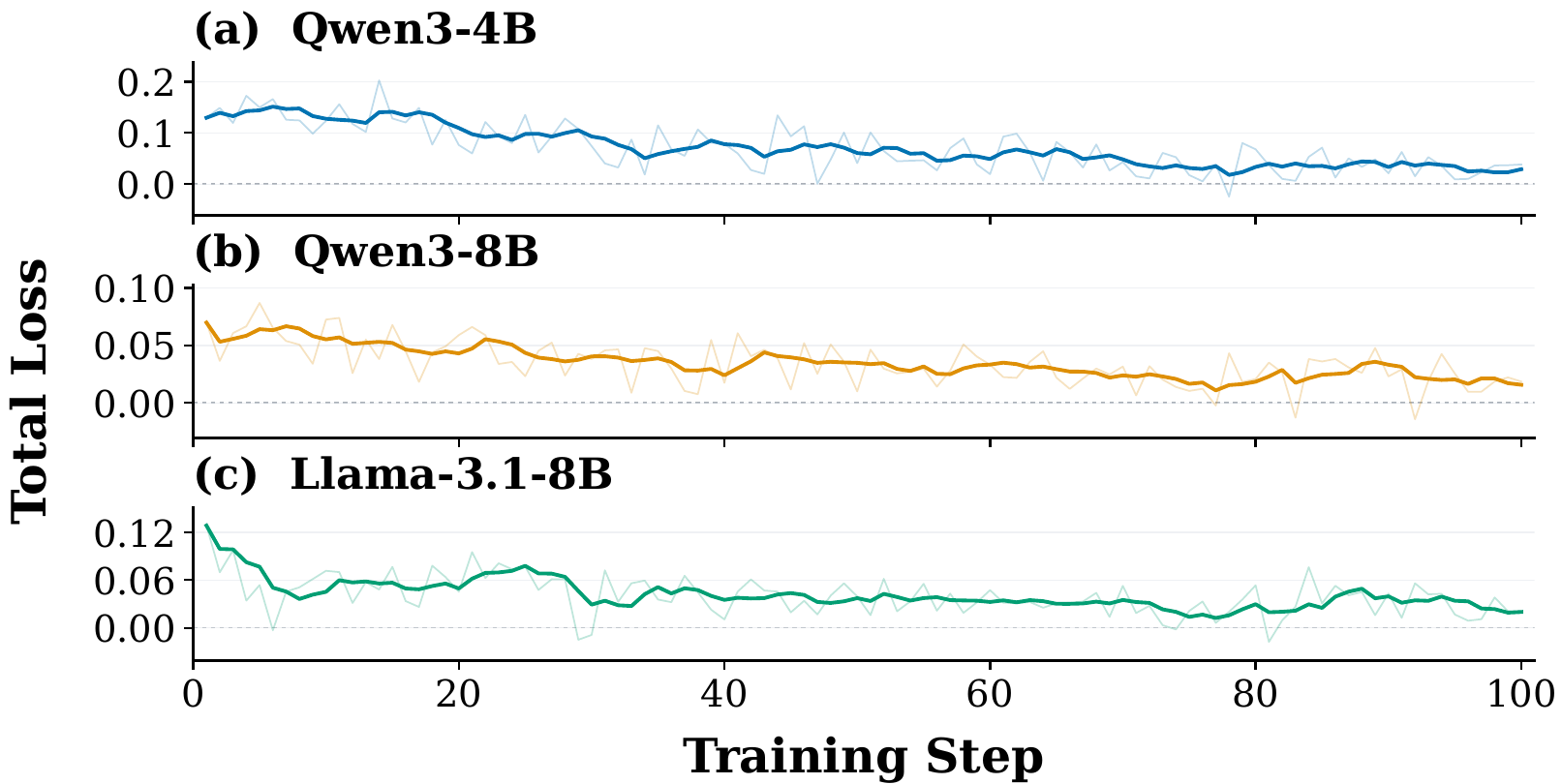}
    \caption{Training losses.}
    \label{fig:losses}
\end{figure}

\subsection{Evaluation}
\label{app:eva-details}
Across all evaluations, we report accuracy averaged over three independent runs, rounding the mean number of correct predictions to the nearest integer before normalizing by the total number of test instances. For baselines without native model selection, we fix the backbone to GLM-5, the model with the largest active parameter in our pool, to ensure strong reasoning capability across tasks. The baselines evaluated in Table~\ref{tab:overall_results} are summarized below:
\begin{itemize}
  \item \textbf{Vanilla LLMs}: Directly use the corresponding LLM to generate responses in a single-turn interaction.
  \item \textbf{RouteLLM}~\cite{ong2025routellm}: A classifier-based binary routing method that chooses between GLM-5 and Qwen3.5-35B-A3B for each query based on the predicted win rate of the stronger model against a fixed threshold.
  \item \textbf{GraphRouter}~\cite{feng2025graphrouter}: Models relations among tasks, queries, and candidate LLMs with a heterogeneous graph, employing a graph neural network to select the suitable model.
  \item \textbf{ICL-Router}~\cite{wang2026iclrouter}: Constructs capability representations of candidate models from historical performance and routes each query to the model predicted to be the best.
  \item \textbf{RouterDC}~\cite{chen2024routerdc}: A dual contrastive learning based routing approach that selects an LLM based on the similarity between the query and candidate capability vectors.
  \item \textbf{Router-R1}~\cite{zhang2025routerr1}: An RL-based agentic routing method where the router actively dispatches sub-tasks to candidate models during reasoning and aggregates returned information to synthesize the final answer.
  \item \textbf{ReCal}~\cite{yu2026recal}: An RL-based agentic routing method following routing behaviors similar to Router-R1~\cite{zhang2025routerr1}, enhancing policy optimization under multi-task and multi-objective settings via reward decomposition and calibration.
  \item \textbf{xRouter}~\cite{qian2025xrouter}: A multi-turn routing method that allows the router to answer directly, invoke candidate models, or select an existing model response as the final answer.
  \item \textbf{MAS-Orchestra}~\cite{ke2026masorchestra}: Encapsulates diverse collaborative reasoning patterns into callable modules coordinated by an orchestrator, using GLM-5 as the LLM backbone.
  \item \textbf{OrchMAS}~\cite{feng2026orchmas}: Dynamically instantiates agent roles and task prompts while advancing problem-solving via runtime agent feedback, adopting GLM-5 as the backbone.
  \item \textbf{DAAO}~\cite{su2025daao}: Dynamically selects reasoning operators and their LLM backbones based on predicted problem difficulty, constructing an adaptive hierarchical collaborative workflow for each query.
  \item \textbf{MasRouter}~\cite{yue2025masrouter}: Employs cascaded controllers to select collaboration modes, agent counts, roles, and LLM backbones, instantiating tailored MAS per query.
  \item \textbf{RCR-Router}~\cite{liu2025rcrrouter}: Allocates shared memory to agents based on task-role relevance, where all agents are backed by GLM-5 and iteratively analyze and answer queries according to a predefined topology.
  \item \textbf{AgentRouter}~\cite{zhang2026agentrouter}: Employs a knowledge graph to characterize relationships among queries, entities, and candidate agents, using a GNN to compute a weighted aggregation of responses across model-strategy combinations.
  \item \textbf{Uno-Orchestra}~\cite{cui2026unoorchestra}: An agentic orchestration method based on SFT and RL that dispatches subtasks to compatible model-primitive pairs, iteratively delegating or synthesizing the answer based on execution feedback.
\end{itemize}

\section{Derivation of Inference Computational Cost}
\label{app:compute-cost}

Measuring computational overhead via commercial API pricing often fails to reflect actual resource consumption, as pricing fluctuates across service providers and over time. For clarity and conciseness, we instead model a standard decoder-only Mixture-of-Experts model $m$ with $L$ layers and hidden dimension $D$, where each layer activates $k$ equally sized routed experts. Let $P_{\mathrm{sh}}$ denote the parameter count of the always-active backbone matrices—including attention projections, routers, and shared experts—and $P_{\mathrm{exp}}$ denote that of a single routed expert. The active matrix parameter count per token is then $A_m=P_{\mathrm{sh}}+LkP_{\mathrm{exp}}$, as unselected experts incur no matrix operations~\cite{abnar2025parameters,jiang2025moecap}.

Counting one multiplication and one addition as one FLOP each, parameter matrix operations incur a dominant cost of $2A_m$ FLOPs per token. Under standard multi-head causal attention, attending to $t$ positions requires an additional $4LDt$ FLOPs for query-key dot products and attention-weighted value aggregation~\cite{wang2025agenttts}. For a model invocation with input length $n^{\mathrm{in}}$, output length $n^{\mathrm{out}}$, and total sequence length $N=n^{\mathrm{in}}+n^{\mathrm{out}}$, let $F_m$ denote the total computational cost and $\mathcal R_m$ capture all remaining operations. Assuming KV caching within each request and counting causal attention pairs, we obtain:
\begin{equation}
\begin{aligned}
    F_m
    &\approx 2A_mN+4LD\sum_{t=1}^{N}t+\mathcal R_m\\
    &=2A_mN+2LDN(N+1)+\mathcal R_m.
\end{aligned}
\label{eq:moe-flops}
\end{equation}

Since parameter matrix operations dominate overall computation (exceeding 95\% of total FLOPs in our experiments), we omit both the attention interaction term and $\mathcal R_m$, yielding the FLOPs proxy $\widehat F_m=2A_m(n^{\mathrm{in}}+n^{\mathrm{out}})$. Taking the active parameter count of the smallest model in our LLM pool, Qwen3.5-35B-A3B ($A_0=3\times10^9$), as the reference baseline, we define $2A_0$ FLOPs as one cost unit. Let $\mathcal I_W(\tau)$ denote the index set of worker calls along trajectory $\tau$, where the $j$-th call invokes model $m_j$ with input length $n_j^{\mathrm{in}}$, output length $n_j^{\mathrm{out}}$, and estimated cost $\widehat F_j$. The normalized trajectory cost is then defined as:
\begin{equation}
\begin{aligned}
    C(\tau)
    &\mathrel{:=} \frac{\sum_{j\in\mathcal I_W(\tau)}\widehat F_j}{2A_0}\\
    &=\sum_{j\in\mathcal I_W(\tau)}
    \frac{A_{m_j}}{A_0}
    \bigl(n_j^{\mathrm{in}}+n_j^{\mathrm{out}}\bigr).
\end{aligned}
\label{eq:cost-trajectory}
\end{equation}

In practice, we approximate $A_m$ using the active parameter counts reported by model providers. For dense models, this equals the total parameter count, since all weights participate in each autoregressive forward pass. Active parameter counts for all evaluated models are summarized in Table~\ref{tab:llm-activation}.

\begin{table}[t]
  \caption{Active parameter counts of the models. For dense models ($^*$), the total parameter count is reported. $^\ddagger$ denotes models used only in Section~\ref{sec:resource-generalization}.}
  \label{tab:llm-activation}
  \centering
  \resizebox{\linewidth}{!}{%
    \begin{tabular}{lc @{\hspace{1.2em}} lc}
      \toprule
      Model & Act. Params (B) & Model & Act. Params (B) \\
      \midrule
      Qwen3.5-35B-A3B          & 3  & DeepSeek-V4-Flash           & 13 \\
      Qwen3-4B$^*$             & 4  & Qwen3.5-397B-A17B           & 17 \\
      Qwen3.8-Flash$^\ddagger$ & 6  & Kimi-K2.8-Preview$^\ddagger$ & 32 \\
      Qwen3-8B$^*$             & 8  & GLM-5                       & 40 \\
      Llama-3.1-8B$^*$         & 8  & Qwen3.8-Max$^\ddagger$      & 95 \\
      Qwen3.5-122B-A10B        & 10 &                             &    \\
      \bottomrule
    \end{tabular}%
  }
\end{table}

\end{document}